# Performance Dynamics and Termination Errors in Reinforcement Learning – A Unifying Perspective

Nikki Lijing Kuang
Department of Computer Science and Engineering
University of California, San Diego
La Jolla, CA, USA
l1kuang@ucsd.edu

Clement H. C. Leung
Centre for Applied Informatics
Victoria University
Melbourne, Victoria, Australia
clement.leung@vu.edu.au

*Abstract—* In reinforcement learning, a decision needs to be made at some point as to whether it is worthwhile to carry on with the learning process or to terminate it. In many such situations, stochastic elements are often present which govern the occurrence of rewards, with the sequential occurrences of positive rewards randomly interleaved with negative rewards. For most practical learners, the learning is considered useful if the number of positive rewards always exceeds the negative ones. A situation that often calls for learning termination is when the number of negative rewards exceeds the number of positive rewards. However, while this seems reasonable, the error of premature termination, whereby termination is enacted along with the conclusion of learning failure despite the positive rewards eventually far outnumber the negative ones, can be significant. In this paper, using combinatorial analysis we study the error probability in wrongly terminating a reinforcement learning activity which undermines the effectiveness of an optimal policy, and we show that the resultant error can be quite high. Whilst we demonstrate mathematically that such errors can never be eliminated, we propose some practical mechanisms that can effectively reduce such errors. Simulation experiments have been carried out, the results of which are in close agreement with our theoretical findings.

*Keywords-Error probability; premature termination; reflection principle; reinforcement learning; random walk; rewards margin; termination condition*

## I. INTRODUCTION

Reinforcement Learning (RL) has been actively applied to address issues occurred in dynamic systems in Artificial Intelligence and Machine Learning areas, especially with recent breakthroughs in deep learning [3] that enable fresh perspectives of looking at the problems [2][10]. In RL, the problem of finding termination conditions and convergence guarantee is a research topic of practical importance that has received extensive attention. Specifically, a decision needs to be made at some point as to whether it is worthwhile to carry on or to terminate the learning process.

The termination condition is deemed to be met in classical RL algorithms when an optimal policy is attained. Among these methodologies designed for small-scale problems with finite states and actions, *Q*-learning is a widely-adopted stochastic approximation algorithm for which proofs showing its convergence under certain conditions are available [4][5][6], and the corresponding convergence rate has been studied in [8]. However, many of these algorithms suffer the common problem of slow convergence. Taking variants of *Q*-learning as an example, the noise in the approximation of value function could significantly slow down the convergence process [17]. Methods such as reducing input dimensionality [10], state space reduction [11][12], learning upper and lower estimates of the value function [9] are designed for performance improvement. A somewhat unsatisfactory situation is that of frequent failures to converge to an optimal solution when applied to practical industrial problems [7]. The fact that the optimum is not always attainable implies such termination condition may result in an infinite learning-loop, inducing inefficiency and futile efforts.

Finding optimal policies in continuous space is even more intricate, where searching for the greedy action in each state is time-consuming. To handle large-scale practical problems in RL systems, online learning algorithms are developed, as the effectiveness of fitting problems with continuous (or even infinite) state and action spaces into traditional RL methods largely depends on function approximation accuracy [13]. Works include the kernel-based temporal-difference learning approach [7], the Cacla algorithm [20] and CMA-ES [19], which are introduced for online policy search. Though existing online RL algorithms are tractable for continuous spaces with improved performance, persistence of excitation condition [14][18] that is difficult to verify has to be satisfied in order to converge to the optimum. In [14], an integral RL algorithm on actor-critic structure utilizes the experience replay to accelerate convergence. Still, current methods are sample inefficient [14][16], i.e., requiring a large number of real samples for learning, and suffers from instability concerning convergence guarantee [16].

Although the problem of termination conditions has been studied by different researchers, most of the available works are algorithm-oriented and cannot be applied universally. The motivation of this paper is to provide a unifying perspective of the termination problem in RL, independent of particular RL algorithms serving different purposes. We analyze the termination error whereby a learning process is prematurely terminated when the first few rewards appear to be unpromising, despite a large number of positive rewards occurring later. For a given RL environment, the conditional probabilities of incorrect termination in a learning episode are studied. Closed-form expressions and error bounds are derived, and mechanisms for limiting such errors are presented. The advantage of our method is that the results are inherently tractable in the RL process, and it is able to be deployed as a useful performance evaluation metric to assess the effectiveness of general RL algorithms.

## II. PREMATURE TERMINATION ERROR PROBABILITY AND ITS REDUCTION

### A. Condition for Terminating a Learning Episode and Bounds on the Termination Error

In a particular learning episode, for example when one tries out a new route to work, it is necessary to eventually decide – by considering the distance and traffic delay in that period – if the route is worth taking or perhaps to give up learning on that particular route entirely. By considering the time duration for the journey, each trial would result in a positive reward $R$ or a negative reward $R'$. Very often, such situations are intrinsically stochastic in nature as the underlying determinants of the rewards are the results of random behavior, which in the foregoing example are the varied traffic patterns. Likewise, in considering a new algorithm for an online recommender system, its effectiveness is decided based on whether the recommended items are taken up by users. The stochastic elements here can be linked to the instance of randomized ranking, the relative degrees of exploitation and exploration in the recommendation mechanism, the momentary mood of the users, their cultural, family and educational backgrounds, as well as the hidden personality profile and preference nuances. Here, we shall consider a model to incorporate such randomness that is applicable to a variety of RL situations.

Let the total number of negative rewards be $N_{R'} = x$, and that of positive ones be $N_R = kx$, with the total number of trials, $N$, in the learning episode being $N = N_{R'} + N_R = x + kx$. In general, $k$ can be any value in the interval $I' = (0, \infty)$, but for the present analysis, we shall use the restricted interval $I = (1, \infty)$, i.e. $k \in I$, corresponding to situations where the number of negative rewards is bounded by the positive ones. In this study, we shall focus on the number of rewards rather than the magnitude of each reward; i.e. all rewards, no matter positive or negative, are assumed to carry equal weight. Since rewards arrive sequentially and randomly – i.e. an observed reward can be positive or negative, depending on the number of positive and negative rewards remaining, with each chosen with equal probability – it is entirely possible that the learner will give up too early before the full picture emerges; for example, if the first few happen to be negative rewards, one may simply give up and not bother to observe the remaining ones which may well be overwhelmingly positive. A rule often adopted by many practical learners is the following:

**Rule I**: *Stop the learning process once the number of cumulative negative rewards is greater than or equal to the number of cumulative positive rewards*

Since $kx > x$, conclusions should be drawn upon completing the full learning episode of $x+kx$ trials, which in our example, is whether one should adopt the new route. Given these, we wish to determine the conditional probability of termination before completing the full learning episode. In particular, we wish to determine the probability that the negative rewards exceed or equal to the positive ones in the sequential process before the learning episode is completed, which we shall refer to as the *termination error*. In particular, if $k \gg 1$, one should never prematurely terminate.

**Theorem 1.** The termination error $p$ under Rule I is given by

$$p = \frac{2}{k+1}. \quad (1)$$

**Proof.** Consider the first reward. Instead of using a sequential sampling without replacement approach, we shall use the simpler approach of considering the different paths in leading to the final destination, where a path corresponds to a particular learning sequence. Such situation may be viewed as a type of one-dimensional *random walk* starting from the origin in which a positive reward corresponds to a step up, while a negative reward corresponds to a step down. The actual displacement therefore corresponds to the *net* number of positive rewards over the number of negative rewards. Thus, for a positive first step, a premature termination would occur on the first return to the origin, while a negative first step would have immediate termination.

Incorporating the time dimension, we let the vertical $y$-axis represent the net number of positive rewards, and the horizontal $t$-axis represent the number of trials (Fig. 1). Each path starts at the point (0, 0); at the end of the full learning episode of $x+kx$ trials, the path will terminate at the point $(x+kx, kx-x)$. Now a complete path that begins with $R'$ (i.e. a step down from (0,0) to $(1, -1)$) occurs with probability

$$\frac{x}{kx+x}. \quad (2)$$

Such a path will surely lead to an undesired premature termination, since $N_R > N_{R'}$ the number of positive rewards will catch up at some point, where the number of negative rewards and the number of positive ones exactly balance, in which case the path will touch the $t$-axis (see Fig. 1).

Next consider a path that begins with $R$ (i.e. a step up from (0,0) to (1,1)) and yet reaches a situation where the number of negative rewards and the number of positive rewards also exactly balance. Such a situation will again cause premature termination, since touching the $t$-axis means that the number of positive rewards and the number of negative ones are the same, requiring termination under Rule I. By the Reflection Principle [1], such a path that begins with $R$ will have a one-to-one correspondence with a path that begins with $R'$ by changing $R$ to $R'$ and $R'$ to $R$ through reflection on the $t$-axis for the section of the paths prior to touching the $t$-axis.

Our required result is the probability of learning sequences that reach a balance at some point, which equals to the probability of the interested sequences with the beginnings of $R'$ or $R$. Thus, our required probability is twice that of the probability given in (2), hence, completing the proof.

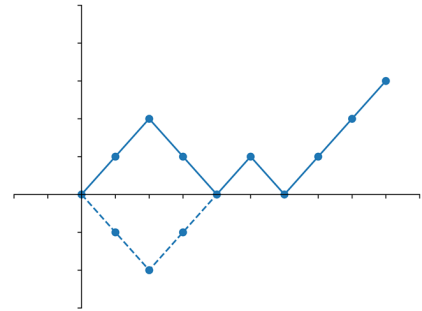

Figure 1.  Path Analysis of Ramdom Walks

$$2 \times \frac{x}{kx + x} = \frac{2}{k+1}.$$

We see that, even when $k = 2$, i.e. the number of positive rewards is twice that of the negative ones, this error is quite high with 67%, and when the positive rewards is five times the negative ones, the error is still considerable with 33%. Very often, we would like to have a bound $M$ ($<1$) on the error. From (1), we have

$$\frac{2}{k+1} \leq M. \qquad (3)$$

This gives

$$\frac{2-M}{M} \leq k. \qquad (4)$$

Fig. 2(a) gives different termination errors for various $k$. We see that for $M$ to be small, $k$ should be large. For example, if we wish to have less than 20% error, then $k$ should be $\geq 9$. This implies that the total number of trials required should be not less than $10x$; thus, for 3 negative rewards, then the number of trials should be greater than 30.

Apart from representing $N_R$ as some multiple of $x$, the rewards margin is often used, i.e. $N_R = N_{R'} + r$, where $r$ is the rewards margin.

**Corollary 1.** In terms of the rewards margin, the termination error $p$ under Rule I is given by

$$p = \frac{2x}{2x + r}. \qquad (5)$$

**Proof.** Using the same argument as in the proof of Theorem 1, the required probability is given by

$$2 \times \frac{x}{x + r + x}.$$

This gives the required result

$$\frac{2x}{2x + r}.$$

To have a bound $M$ on the error, from (5), we have

$$\frac{2x}{2x + r} \leq M, \qquad (6)$$

giving

$$\frac{2(1-M)}{M} \leq \frac{r}{x}, \qquad (7)$$

where $r/x$ is the proportional rewards margin. Fig. 2(b) plots the proportional rewards margin against different termination errors, and Fig. 3 plots $r$ against different termination errors for different values of $x$. We observe that for relatively small errors, as $N_{R'}$ increases, the rewards margin increases significantly, while when the errors become large, such increase becomes less dramatic. For example, for an error of 20%, $r = 8x$, while an error of 50%, we have $r = 2x$.

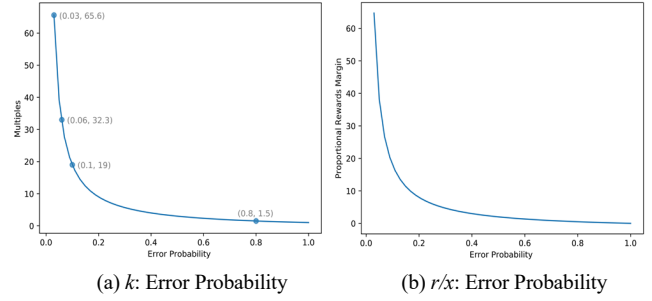

(a) $k$: Error Probability  (b) $r/x$: Error Probability

Figure 2. Analysis of Termination Errors and Bounds

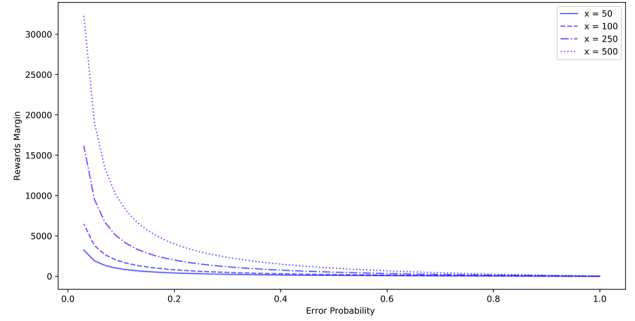

Figure 3. Rewards Margin against Error Probability

### B. Reducing the Conditional Termination Error

One possibility that leads to premature termination is when the sequence of reinforcements begins with $R'$, in which the number of negative rewards exceeds that of positive ones immediately. This occurs with probability $x/(kx+x) = 1/(k+1)$, and this probability can be considerable especially for relatively small $k$. By modifying Rule I to allow the learning process to continue in spite of receiving an initial negative reward, the termination error can be significantly reduced for such situation. The modified rule is:

**Rule II**: *When the first reward is a negative reward, continue to observe the second reward. If the second reward is positive, continue but subsequently stop the learning process once the number of cumulative negative rewards is greater than or equal to the number of cumulative positive rewards; otherwise stop the learning process immediately.*

To analyze this situation, we ignore the first negative reward and start from the second trial, resulting in a situation where we have $kx+x-1$ trials remaining, with $kx$ positive rewards and $(x-1)$ negative ones among them.

**Theorem 2.** The termination error under Rule II is

$$p' = \frac{2(x-1)}{kx + x - 1}. \qquad (8)$$

**Proof.** From Theorem 1, changing the number of negative rewards from $x$ to $x$-1 using (2), we obtain

$$\frac{x-1}{kx + x - 1}.$$

Applying the reflection principle and following similar argument as in the proof of Theorem 1, we obtain for our required probability, hence, completing the proof.

$$\frac{2(x-1)}{kx+x-1}.$$

**Corollary 2.** The termination error under Rule II is strictly less than that of Rule I; i.e. $p' < p$, and that the magnitude of error reduction is

$$\frac{2k}{(k+1)[(k+1)x-1]}. \quad (9)$$

**Proof.** We wish to establish

$$p' = \frac{2(x-1)}{kx+x-1} < \frac{2}{k+1} = p.$$

This is equivalent to

$$(k+1)(x-1) < (k+1)x - 1 \Leftrightarrow 1 < k+1,$$

which is obviously true since $k > 1$.
Next, the required result can be obtained from

$$p - p' = \frac{2}{k+1} - \frac{2(x-1)}{kx+x-1} = \frac{2k}{(k+1)[(k+1)x-1]},$$

Alternatively, in terms of the rewards margin, we have

$$p' = \frac{2(x-1)}{2x+r-1}, \quad (10)$$

with corresponding error reduction

$$\frac{2(x+r)}{(2x+r)(2x+r-1)}. \quad (11)$$

### III. EXPERIMENTATION AND RESULTS INTERPRETATION

To compare and validate theoretical results, simulation experiments are designed and performed for both rules. A process of learning is triggered by randomly selecting rewards, which terminates per Rule I and Rule II respectively. If the learning carries on until the entire counting process finishes, it is deemed to be a successful learning episode.

Table I summarizes the results of termination error under Rule I and and Rule II, where Pos is used to signify the number of positive rewards, and Neg, the number of negative rewards. In each scenario, twenty parallel tests are executed for both rules, within which 30,000 learning processes are included. Average Simulated Termination Error is obtained from the total twenty parallel tests, where the error for a single test equals to the ratio of the number of failure learnings to the number of total learnings. Standard deviation STD and absolute error are given for error measurement. The absolute error is calculated as the absolute difference between the simulated termination error and the theoretical termination error, and standard deviation of simulation results in twenty tests is used to measure the simulation steadiness. At the end of Table I, $p - p'$ is measured both theoretically and experimentally.

TABLE I. SIMULATION RESULTS

| Simulation Scenario | 45 Pos 30 Neg | 80 Pos 20 Neg | 90 Pos 10 Neg | 190 Pos 10 Neg | 760 Pos 40 Neg | 650 Pos 10 Neg |
|---|---|---|---|---|---|---|
| $k$ | 1.5 | 4 | 9 | 19 | 19 | 65 |
| $r$ | 15 | 60 | 80 | 180 | 720 | 640 |
| $r/x$ | 0.5 | 3 | 8 | 18 | 18 | 64 |
| Theoretical $p_t$ | 0.8 | 0.4 | 0.2 | 0.1 | 0.1 | 0.0303 |
| Avg Simulated $p_s$ | 0.8003 | 0.4004 | 0.2002 | 0.1001 | 0.0998 | 0.0303 |
| STD ($p_s$) | 0.0018 | 0.0033 | 0.0019 | 0.0019 | 0.0016 | 0.001 |
| Abs Error $|p_t - p_s|$ | 0.0003 | 0.0004 | 0.002 | 0.0001 | 0.0002 | 0.000 |
| Theoretical $p_t'$ | 0.7838 | 0.3838 | 0.1818 | 0.0905 | 0.0976 | 0.0273 |
| Avg Simulated $p_s'$ | 0.7835 | 0.3841 | 0.1820 | 0.0896 | 0.0966 | 0.0291 |
| STD ($p_s'$) | 0.0030 | 0.0072 | 0.0079 | 0.0063 | 0.0070 | 0.0077 |
| Abs Error $|p_t' - p_s'|$ | 0.0003 | 0.0003 | 0.0002 | 0.0009 | 0.0010 | 0.0018 |
| $p_t - p_t'$ | 0.0162 | 0.0162 | 0.0182 | 0.0095 | 0.0024 | 0.0030 |
| $p_s - p_s'$ | 0.0168 | 0.0163 | 0.0182 | 0.0105 | 0.0032 | 0.0012 |

It is easily seen that in each scenario, by comparing $p_s$ with $p_t$, $p_s'$ with $p_t'$, the simulated termination error is very close to the theoretical one under both rules. The low STD suggests the termination error tends to be stable in each parallel test. The negligible absolute errors between $p_s$ and $p_t$, $p_s'$ and $p_t'$, corroborate the correctness of our theoretical results. It is also interesting to observe that for Rule I, when $p_t$ is large (0.8), the number of positive rewards is 1.5 times of the negative ones, while the proportion multiple $k = 4$ is able to decrease it to 0.4; and $p_t$ even goes down to 0.2 with a value of $k < 10$. This could be inspiring in the sense that a small increase of $k$ is able to restrict the error probability to 20%. Similar statements also hold true for Rule II. The two scenarios with different total rewards but the same value of $k$ and the same $r/x$ results in the same $p_t$, demonstrating that $p_t$ does not depend on specific values of rewards. Furthermore, with $p_t - p_t'$ and $p_s - p_s'$, we see that the termination error under Rule II is strictly less than that of Rule I.

With our study here, one can easily analyze the problem of convergence guarantee. Suppose $k = 4$, the expected termination error is 40% under Rule I. We then easily learn that when Rule I is adopted, although an optimal policy exists for the practical problem, there is 40% chance that the learning terminates in advance because of the unpromising rewards received in the process. To ensure the convergence guarantee, the algorithm needs to be modified to limit the influence of early negative rewards as suggested in the previous section. As can be seen, when analyzing the effectiveness of RL algorithms, our study provides a useful evaluation metric.

## IV. CONCLUSIONS AND FUTURE WORK

In many RL situations, stochastic elements are often involved in governing the occurrence of rewards, with sequential occurrences of positive rewards randomly interleaved with negative rewards. While the overall positive rewards can be much greater than the negative ones, the occurrence of a cluster of negative rewards at the initial stages of the learning episode can cause premature and erroneous termination of the learning process. Such situations are analyzed and we have found the termination errors can be quite high even when the overall number of positive rewards is many times that of the negative rewards. Closed-form expressions of such errors are obtained using combinatorial analysis, and mechanisms for substantially reducing such errors are also given.

We have mathematically demonstrated that termination error in RL can never be eliminated, since $2/(k+1)$ is always positive for any finite $k$. Even when $k \gg 1$ or when $r \gg x$, where an optimal policy exists to conclude learning success, there is still the possibility of premature termination to conclude learning failure. In fact, the correctness of learning may be viewed as a binary classification problem with the labels *success* and *failure*, and the optimal policy is to minimize the classification error. Using ROC analysis [15], the optimality of different policies may be evaluated and compared. As in any classification problem, in general neither Type I nor Type II errors can be avoided. In this paper, we concentrate primarily on quantifying Type II errors. It will be useful in future studies that address Type I errors as well for different assumptions and operating conditions. In addition, whilst we have proposed a mechanism for bounding the termination error, it is useful to investigate additional mechanisms for doing so which will further limit such errors and at the same time achieve greater fault tolerance.